# Continuous Sign Language Recognition System using Deep Learning with MediaPipe Holistic

Sharvani Srivastava[1], Sudhakar Singh[2], Pooja[3], Shiv Prakash[4]

---------------------------------------------

Sharvani Srivastava
sharvanisri28@gmail.com

Sudhakar Singh
sudhakar@allduniv.ac.in
Corresponding Author, ORCID: 0000-0002-0710-924X

Pooja
cs.pooja@allduniv.ac.in

Shiv Prakash
shivprakash@allduniv.ac.in

[1,2,3,4]Department of Electronics and Communication, University of Allahabad, Prayagraj, India

*Abstract –* Sign languages are the language of hearing-impaired people who use visuals like the hand, facial, and body movements for communication. There are different signs and gestures representing alphabets, words, and phrases. Nowadays approximately 300 sign languages are being practiced worldwide such as American Sign Language (ASL), Chinese Sign Language (CSL), Indian Sign Language (ISL), and many more. Sign languages are dependent on the vocal language of a place. Unlike vocal or spoken languages, there are no helping words in sign language like is, am, are, was, were, will, be, etc. As only a limited population is well-versed in sign language, this lack of familiarity of sign language hinders hearing-impaired people from communicating freely and easily with everyone. This issue can be addressed by a sign language recognition (SLR) system which has the capability to translate the sign language into vocal language. In this paper, a continuous SLR system is proposed using a deep learning model employing Long Short-Term Memory (LSTM), trained and tested on an ISL primary dataset. This dataset is created using MediaPipe Holistic pipeline for tracking face, hand, and body movements and collecting landmarks. The system recognizes the signs and gestures in real-time with 88.23% accuracy.

## I.  INTRODUCTION

Communication is a vital part of our life. Talking to someone, listening to the radio or speeches, watching television, reading books, magazines, or newspapers, and writing letters, etc. are all examples of communication. Without communication, a person will be leading to a lonely monotonous life just performing only his own tasks like a robot. As seen by these examples, there are several categories of communication such as formal and informal kinds of communication, grapevine communication, written and oral communication (oral can be face-to-face and distance), non-verbal communication, visual communication, active listening communication, and feedback communication [1]. The formal type of communication (well known as official communication) is operated through pre-determined channels. The Grapevine communication (also known as unofficial communication) occurs between individuals within one's profession which may involve clients, colleagues, supervisors, employers, etc. Word of mouth information is considered to be the main feature of grapevine communication as well as spontaneity and free-flow without any formal protocols or constructs. It occurs in the external and internal informal channels leading to effective operation and benefit of an organization. Face-to-face oral communication is considered to be the most recognized and common form of communication. In this kind of communication, the main aim is to express directly to others through words. It may be formal or informal communication. Oral communication (distance) occurs between people at a distance using different methods and technologies like mobile phones, 2-way webinars, video-conferencing, Voice over Internet Protocol (VoIP), and so forth. Written communication, as per its name, is communication taking place in written form via different means such as letters, notices, emails, messages, advertisements, etc. Feedback communication is where individuals in leadership positions i.e. instructors, heads, supervisors, employers, directors, and so forth in different types of organizations (like educational institutions), implement numerous assessment methods to evaluate an individual's performance and give feedback in respect of their performance. Visual communication is a common class of communication in people's daily lives. Televisions, social networking, smartphones, etc. are the main sources of visual communication. Active listening is considered as a type of communication that has the utmost significance. It is a significant characteristic which is needed to be practiced to make communication meaningful and effective. Non-verbal is a subtle type of communication that uses facial expressions, eye contact, body language, gestures, space, touch, and the personality of the individual [1].

The use of facial expressions, movements, body language, etc. in non-verbal communication is of great help to people with hearing impairment as there is nothing spoken or heard, the whole communication depends upon observing the person they are communicating with. Deaf is the disability which impairs a person's hearing making him unable to hear and dumb is the disability which impairs a person's speaking making him incapable to speak. These disabilities hinder them to communicate freely with other people. To help them communicate easily, sign language comes into role enabling a person to communicate with others without any word. But even with sign language, there still lies a problem that not many people are well-versed with sign language limiting deaf and dumb people to not being able to communicate with everyone. Here, technology plays an important role. With the help of a technology-driven solution, sign language gestures can be translated into spoken language and vice-versa.

There are numerous sign languages used world-wide. It depends upon the spoken language of a particular place. Indian Sign Language (ISL) is used in India by hearing impaired persons. In ISL, the gestures are made by either single hand or double hands which leads to two different gestures expressing the same thing. The use of single hand and double hands increases the complexity of ISL. Globally, the field of SLR has conducted extensive studies on various sign languages such as ISL and ASL [2][3][4].

Various machine learning and deep learning based models and methods have been employed to develop SLR systems. Some commonly observed methods are K-Nearest Neighbour, Artificial Neural Network, Support Vector Machine, Hidden Markov Model, and Convolutional Neural Network abbreviated respectively as KNN, ANN, SVM, HMM, and CNN [5][6][7][8][9][10]. Traditional neural networks operate only on the current information provided and do not take any previous information into consideration which may be essential to derive the desired output in some cases. This issue is addressed in the Recurrent Neural Network (RNN) [11] where it has the ability to consider the previous information to derive the output. But the limitation is that it works only with short-term dependencies. It completely modifies the existing information to add new information. This is when Long Short-Term Memory (LSTM) comes into the role to address the limitation of RNN. LSTM, introduced by Hochreiter & Schmidhuber [12] in 1997, is a special type of RNN having the capability of learning long-term dependencies. People widely use LSTM due to its exceptional performance on a wide range of problems. The main goal of LSTM is to avoid long-term dependency problem. It has the ability to modify information selectively to produce better results. Mittal et al. [13] developed a continuous SLR using the modified-LSTM model. They used a Leap motion sensor for acquiring the sign inputs. They proposed a modified LSTM classifier which is used for recognising the continuous signed sentences with the help of sign sub-units and achieved 72.3% accuracy for the signed sentences and 89.5% for the isolated sign words.

The present paper develops a continuous SLR system using MediaPipe Holistic [14] and LSTM. MediaPipe Holistic [15], [16] helps in tracking face, hands, and pose without any difficulty as well as extracting and collecting the key points. Differentiating the body from the background also becomes a simple task due to its use. Because of the dataset being created using MediaPipe Holistic, the neural network has only about half a million parameters as opposed to millions of parameters that are usually observed in such networks and it is faster to recognize a sign shown to the system.

After the introduction section, this paper is arranged as follows. Section II discusses the review of literature on the state-of-the-art in this filed. Section III comprises data acquisition while Section IV provides the details of the methodology used to develop this system. Section V contains the findings found and finally, Section VI summarises the paper as a conclusion.

## II. RELATED WORK

Sign language is defined as a meaningful structured set of the hand gestures which is used by hearing impaired people for communicating in daily life [17]. It depends upon the visuals perceived by a person on watching hand, facial, and body movements of others. The number of sign languages used all over the world exceeds 300 [18]. Even with an abundance of sign languages, the part of the population that retains the knowledge and awareness of sign language is very small. Due to this lack in skill of any sign language, it is difficult for hearing impaired people to talk to the normal people freely. This is where SLR provides a solution for everyone to communicate easily without being well-versed with sign language. An SLR system is used to identify the movements of sign language and then convert them into the vocal language like English language.

The SLR is a field with great potential for research. A number of research studies have been performed in this field but there are still shortcomings and opportunities that need to be addressed. The SLR systems are of two kinds – isolated as well as continuous SLR. An isolated system is designed and developed in a way that it is capable of recognizing isolated signs i.e., it is able to recognize only individual signs. A continuous SLR system has the ability to recognize and translate the whole sentence instead of individual signs. Continuous SLR is built by using isolated SLR systems as the building blocks [11][19].

Data acquisition is a prerequisite and integral phase of SLR. In the literature, different methods have been used for data acquisition which can be categorised into two types: vision-based methods and direct measurement methods [17][20]. The use of motion capturing systems, motion data gloves or sensors can be seen in the direct measurement methods. The extracted motion is able to track accurately the hand, finger, and other parts of the body leading to the robust SLR techniques outcome. The vision-based methods of SLR depend upon the mining of discriminative temporal and spatial from RGB images. The strive to track and extract the hand regions initially prior to classifying them into gestures can be observed in most of the vision-based approaches [17]. As skin color is distinguishable, it helps in hand detection by the semantic segmentation and detection of skin color [21][22]. It is easy for systems to incorrectly recognize other body parts such as arms and face as hands, this led to the development of recent detection methods where identification of only the moving parts depends on detection and subtraction of face and background [23][24]. Filtering techniques like Kalman and particle filters have been employed by authors for achieving accurate and robust hand tracking. This can be observed especially in cases of occlusions [23][6].

Different devices have been used for the purpose of data acquisition. The commonly used device in SLR for the input process is the camera [25]. Apart from the camera, other devices have also been used like Microsoft Kinect. It provides depth video stream and color video stream simultaneously. The depth data is helpful in background segmentation. Not only the devices but other methods have also been employed for data acquisition like accelerometers and sensory gloves. One of the sensors employed for data acquisition is a touchless controller called Leap Motion Controller (LMC) which is able to operate around 200 frames per second and has the capability of detecting and tracking fingers, hands, and finger-like objects [2][26][13][20].

Many approaches have been explored for developing SLR systems – isolated SLR as well as continuous SLR using HMM [6][7][8], LSTM [27][13], RNN [11], CNN [9][10], Conditional Random Fields (CRF) [28], Deep Neural Network [9], Residual Network [29], etc. Cheng et al. [3] are the first to propose and employ a fully convolutional neural network for continuous SLR. A systematic model has been proposed that modalised sign language as glosses in place of sentences. Their fully convolutional network architecture achieved improvement in complexity and recognising the real-world situation. Pu et al. [30] designed an alignment network which optimises the weakly supervised continuous SLR. Their framework included a 3D residual network (3D ResNet) for the feature extraction. They contributed a unified deep learning architecture integrating the encoder-decoder network and Connectionist Temporal Classification (CTC) for the continuous SLR, a soft dynamic time warping alignment constraint between LSTM and CTC decoders, an iterative optimization strategy for training the encoder-decoder network, and a feature extractor alternately with the alignment proposals by warping paths.

Similar to the above-mentioned works, research has been performed in the field of SLR for ISL as well. Classification of single-handed as well as double-handed ISL recognition has been performed using machine learning algorithms in MATLAB by Dutta and Bellary [5]. The authors applied two algorithms i.e., KNN and Back Propagation for training their system and observed that the accuracy of their system is 93-96%. Hore et al. [4] proposed three techniques to recognize ISL gestures, where a dataset consisting of 22 ISL gestures is used. They trained the neural network using a Genetic Algorithm, Evolutionary Algorithm, and Particle Swarm Algorithm that enabled them to achieve 99.96% accuracy. A multimodal framework for an SLR system is proposed by Kumar et al. [31] employing Leap Motion and Kinect. Leap motion was used for capturing movements of hands and fingers while Kinect was used for capturing facial expressions. They contributed a feature-independent classifier combination which makes use of the signals captured from two modalities. A comprehensive experimental evaluation of their approach against the single modality systems has been carried out. They achieved 96.05% accuracy in single hand gesture recognition and 94.27% in double hands gesture recognition.

The study of literature reveals that there are many challenges in the development of SLR systems. Some of them are mentioned as below [28][5].

• It is difficult to extract features from a complex background.

• There is variation in recognition depending upon the signer.

• Sometimes there is an overlapping of hands and face which hinders the recognition.

• There is a lack of universal sign language. Sign language varies in places according to the spoken language.

• Vocabulary is vast which makes it difficult to include many signs in the dataset.

• In ISL, there are many gestures that are made by single hands as well as double hands due to which two different gestures represent a single word or phrase.

This paper addresses the first challenge as listed above. In this paper, the use of MediaPipe Holistic for the purpose of data acquisition helps in easily tracking faces, hands, and poses and extracting and collecting the key points. Its potentiality makes it easy to distinguish the body from the background.

# III. DATA ACQUISITION

In this paper, a real-time continuous sign language recognition (CSLR) system has been developed for the ISL. For action recognition, MediaPipe Holistic has been used to track face, hands, and pose components. Python and OpenCV have been used for data acquisition. The videos were captured for each gesture, in a total of 45 gestures. From the videos, key points were extracted to acquire the data for training.

The dataset has been created using 45 gestures of ISL which were selected for this dataset. 26 of them are the signs for alphabets of the English Language [32] and the remaining 19 represent different words and phrases of ISL as shown in Table 1.

Table 1: Dataset of 45 signs and gestures of ISL

| Alphabets | Words and Phrases |
|---|---|
| A, B, C, D, E, F, G, H, I, J, K, L, M, N, O, P, Q, R, S, T, U, V, W, X, Y, Z | Namaste, Hello, Bye-Bye, Do not understand, Good Afternoon, Good Morning, How are you?, I am fine, My name is, I/Me, India/Indian, Sign, Language, Understand, No, Yes, Sorry, Thank you, Welcome |

For data acquisition, Open CV and Python have been used. Dependencies like cv2, NumPy, os, time, and MediaPipe were imported. OpenCV i.e., cv2 is an open-source library of computer vision and machine learning algorithms. It provides a common framework for computer vision based applications and expedites the usage of machine perceptivity in commercial products. It contains more than 2500 efficient algorithms that comprehensively include both classic and state-of-the-art algorithms of computer vision and machine learning [33][20]. NumPy enables numerical computing with Python. It was used for working with the extracted key points. The dependency os helps in working with file paths. It belongs to Python's standard utility modules providing functions for interaction with operating systems. Python's time module helps in representing time in various ways in code e.g. numbers, strings, and objects and it can also be used in measuring code efficiency or waiting during the execution of code. It helped in adding breaks between capturing videos. MediaPipe offers the cross-platform, customized and adaptable machine learning solutions for the live and streaming media [14]. This dependency was imported using MediaPipe Holistic which provides the live perception and insight of simultaneous face landmarks, human pose, and hand tracking in real-time [15].

For collecting the data, 45 directories were created representing the 45 signs and gestures of the alphabets, words, and phrases. For a single alphabet, word, or phrase, 30 videos of 30 frames were captured. So, in a directory representing one of the 45 signs, 30 sub-directories were created for storing video data and similarly for other directories as well. MediaPipe Holistic provides landmarks for face, pose, and hands. Pose landmarks, face landmarks, right-hand and left-hand landmarks have been used here. A sign captured in a frame is represented by a set of these landmarks which were stored in NumPy arrays. Therefore, in a folder for a video, 30 NumPy arrays were stored each representing a frame by extracting the key points. This process was performed for all 45 signs which resulted in a total of 1350 videos being captured.

# IV. METHODOLOGY USED

Real-time text or spoken language translation from continuous sign language gestures is the goal of the algorithm for a continuous SLR system. A strong foundation for such a system can be provided by combining MediaPipe Holistic with Deep LSTM networks. The steps of the algorithm used for the study is described in Table 2.

Table 2: Algorithmic steps of the methodology

| Sl. No. | Steps | Description |
|---|---|---|
| 1. | Gathering and pre-processing of data | Video Input: Gather endless videos of sign language.<br>Landmark Detection: To identify landmarks on the torso, hand, and face in each frame, use MediaPipe Holistic.<br>Normalization: To accommodate changes in the signer's location and distance from the camera, normalize landmarks to a standard scale and position.<br>Sequence Creation: For every video, create a sequence of normalized landmarks. |
| 2. | Feature Extraction | Take measurements of critical points' distances, velocities, and joint angles, among other attributes, from the landmarks.<br>Create temporal sequences by accumulating features over time. |
| 3. | Architecture Model | Layer of Input: Sequences of features or landmarks entered.<br>LSTM Layers: To capture temporal interdependence, use multiple LSTM layers.<br>Divide the dataset into sets for validation and training.<br>Utilizing lists of landmarks and their corresponding labels, train the model.<br>To prevent overfitting, keep an eye on validation accuracy. |
| 4. | Inference | Record video or use live video input.<br>Utilizing MediaPipe Holistic, extract landmarks.<br>Preprocess and normalize landmarks in real time.<br>Feed the learned LSTM model with sequences.<br>In order to produce spoken or written language, decode the model's output. |
| 5. | Performance Analysis | The suggested model is used to build the confusion matrix and then compute the accuracy and loss during the training of model. |

The main aim is to develop a continuous SLR system for an ISL dataset. The ISL dataset has been captured using OpenCV and Python following the process described in Section III. After the data acquisition process, a label map was created that represents all things (objects) inside the model, specifically, it holds the label of each sign and gesture (alphabets, words, and phrases) along with their respective ID. There are 45 labels in the label map, where each label representing either an alphabet or a word, or a phrase. A unique id has been assigned to each label ranging from 0 to 44. This label map acts as a reference for looking up the class name.

The data that was collected, was then appended into a single array. For this, two arrays were used, one was used for storing all the NumPy arrays that represented the frames of a video and the second array was used to store all the labels i.e., all the signs the videos represent were stored in it. This data was then split into training as well as testing data in the proportion of 95:5. 95% of the acquired dataset was used to train the model while the remaining 5% of dataset was used for testing purposes. Once the data is ready, the next step is to develop the model. The sequential model is imported from the library tensorflow.keras.models which helps in building a sequential neural network, while LSTM and Dense are imported from the library tensorflow.keras.layers. The LSTM layer provides a temporal component to build the neural network allowing action detection whereas the dense layer has a large number of connections with every single neuron receiving the input from each neuron from the prior layer. TensorBoard is imported from tensorflow.keras.callbacks, that helps in tracing and monitoring the model during training for developing the model.

The model is developed using the sequential model to which the neural network layers are added. The neural network has six layers comprising three layers of LSTM and three Dense layers. The first three layers are the LSTM layers followed by the three layers of Dense. The first LSTM layer has 64 units [34]; return_sequence is configured to 'True'; activation function is set to 'relu' i.e. rectifier activation function [35]; and the input_shape is configured as (30,1662) [36]. The second layer of LSTM has 128 units [34]; return_sequence is configured to 'True'; and the activation function is set to 'relu' [35]. The third LSTM layer has 64 units [34]; return_sequence is set to 'False', and the activation function is set to 'relu' [35]. The fourth layer of the neural network i.e., the first Dense layer has 64 units [37]; and the activation function is configured to 'relu' [35]. The second layer of Dense has 32 units [37]; and the activation function is configured to'relu' [35]. The last layer of the neural network i.e., the third Dense layer's unit has been set to actions.shape which is equal to 45 (number of signs and gestures); and the softmax activation function [38], [39] is used and set as 'softmax'.

For compiling the model, Adamax optimizer [40] and categorical_crossentropy loss function [41][42] have been applied and the metrics have been specified as categorical_accuracy [43] to track the accuracy during the model's training. The model is then trained on the training dataset for which epochs are set to 300 [44]. The accuracy achieved by the model and the loss faced by it during the training is mentioned in the subsequent Section V. Once the model has been trained, it is ready for testing and use in real-time.

## V. EXPERIMENTAL EVALUATION

*A. Dataset and Experimental Setup*

The continuous SLR system developed has been trained on an ISL dataset comprising 45 signs and gestures including English Language alphabets, words, and phrases as shown in Table 1. This primary dataset is generated by following the procedure of data acquisition as described in the Section III.

The system has been developed and tested on a machine running Windows 10 operating system and having an intel i5 7[th] generation 2.70 GHz processor, 8 GB of RAM, and a webcam. The webcam has HP TrueVision HD camera with 0.31 Mega Pixel and 640x480 resolution. The programming environment used for the development includes Jupyter Notebook, Python – 3.7.3, and OpenCV – 4.5.5.

*B. Results and Discussion*

The proposed system has been tested using testing data as well as in real-time. After data acquisition, the dataset was split into training and testing dataset in the ratio of 95:5 i.e., 95% of the acquired dataset was used to train the model and the rest 5% of the acquired dataset was reserved for the testing purpose. The model has been tested using this testing data to find the accuracy of the developed SLR system.

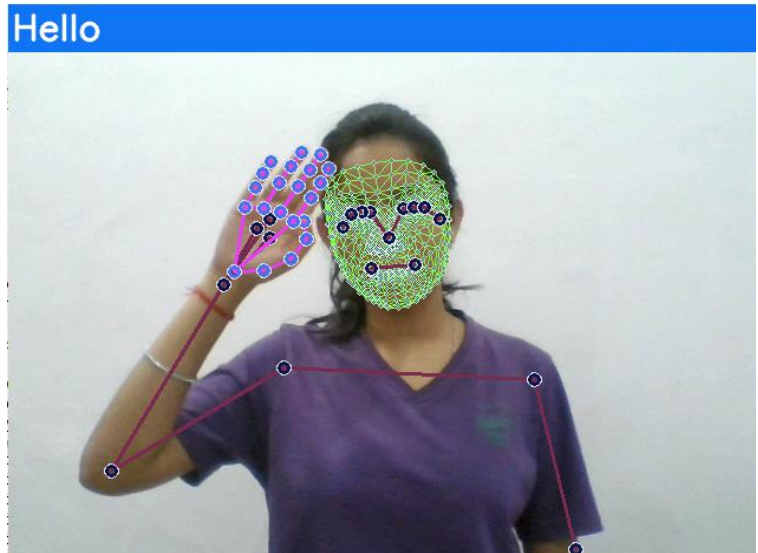
Figure 1: Screenshot of testing in real-time showing the gesture "Hello"

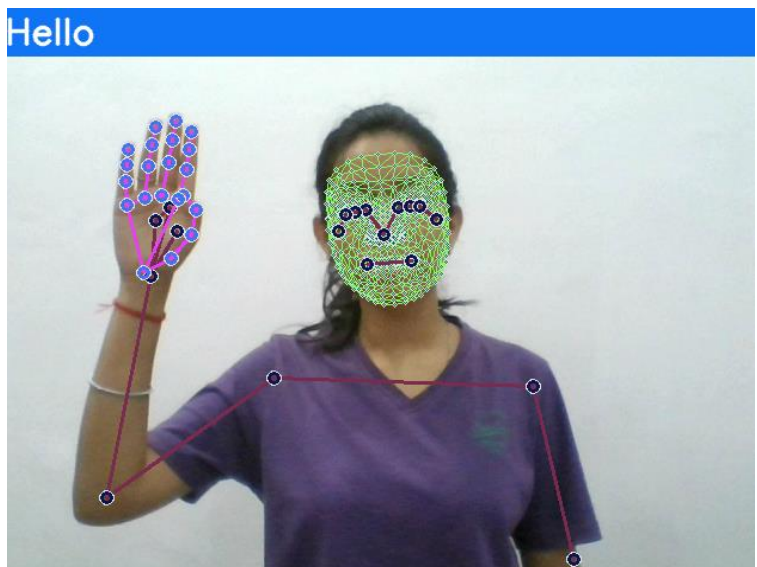
Figure 2: Another screenshot of testing in real-time showing the gesture "Hello"

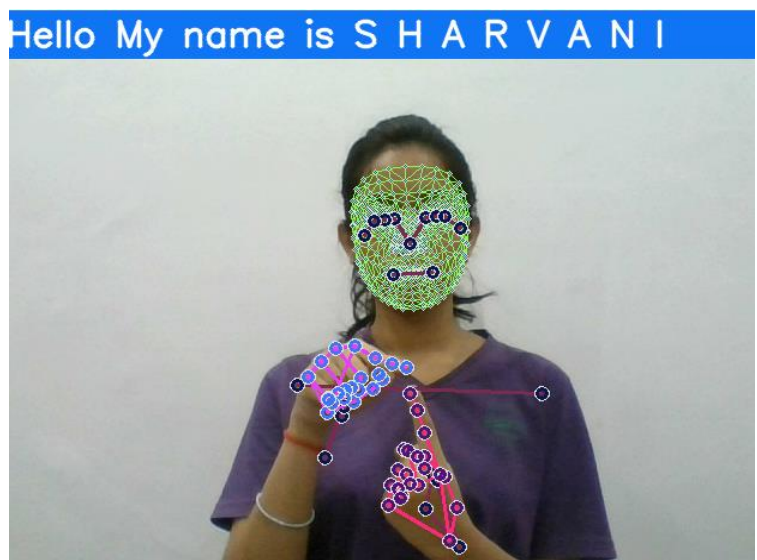
Figure 3: Screenshot of testing in real-time showing the letter "I"

The proposed system has also been tested in real-time where it recognizes the signs and gestures shown to it and the result is displayed on the top to form the whole sentence as shown in Figures 1, 2, and 3.

While training the model, categorical accuracy raised to 1.0 (100%) and loss decreased to 0.0024 as shown in Figure 4.

```
41/41 [==============================] - 4s 104ms/step - loss: 0.0089 - categorical_accuracy: 1.0000
Epoch 293/300
41/41 [==============================] - 4s 104ms/step - loss: 0.0057 - categorical_accuracy: 0.9999
Epoch 294/300
41/41 [==============================] - 5s 110ms/step - loss: 0.0043 - categorical_accuracy: 1.0000
Epoch 295/300
41/41 [==============================] - 4s 105ms/step - loss: 0.0044 - categorical_accuracy: 1.0000
Epoch 296/300
41/41 [==============================] - 4s 105ms/step - loss: 0.0031 - categorical_accuracy: 1.0000
Epoch 297/300
41/41 [==============================] - 4s 107ms/step - loss: 0.0027 - categorical_accuracy: 1.0000
Epoch 298/300
41/41 [==============================] - 4s 105ms/step - loss: 0.0028 - categorical_accuracy: 1.0000
Epoch 299/300
41/41 [==============================] - 4s 105ms/step - loss: 0.0029 - categorical_accuracy: 1.0000
Epoch 300/300
41/41 [==============================] - 5s 117ms/step - loss: 0.0024 - categorical_accuracy: 1.0000
Out[18]: <tensorflow.python.keras.callbacks.History at 0x1679fd145c0>
```

Figure 4: Accuracy and loss during training the model

The accuracy of the system tested using the testing data is found to be 88.23%. The system correctly recognizes signs and gestures shown to it in real-time and displays the sentence on the top.

Due to the use of MediaPipe Holistic, higher accuracy has been achieved with a small amount of data. The use of MediaPipe Holistic followed by the LSTM model has made the designed neural network simple i.e., instead of millions of parameters that neural networks normally have, this model only has about half a million parameters, which is faster to train and efficient in detection in real-time.

## VI. CONCLUSION AND FUTURE WORKS

In this paper, a continuous sign language recognition system has been developed using MediaPipe Holistic and LSTM deep learning model. An ISL dataset of 45 signs and gestures including 26 English Language alphabets and 19 words and phrases has been created. Data acquisition has been performed using OpenCV and Python by webcam. MediaPipe Holistic has been used to collect landmarks of signs in each frame. A neural network of six layers with three LSTM layers followed by three Dense layers has been designed. The model has been trained on 95% of the collected dataset. The system has been tested using a testing dataset which is 5% of the collected dataset as well as in real-time too. The system has been observed to achieve an accuracy of 88.23%.

By using MediaPipe Holistic, sufficiently high accuracy has been achieved even on a small dataset. The neural network is simple because of the use of MediaPipe Holistic followed by the LSTM model, this is why the model is faster to train and able to detect in the real-time.

In the future, more gestures can be added to make a larger dataset. The quality of the dataset can be increased to achieve higher accuracy. The developed system can also be implemented for the other sign languages as well.

**APPENDIX: LIST OF ACRONYMS**

| Acronym | Full Name |
|---|---|
| ADB | Ada Boost Algorithm |
| CSL | Chinese Sign Language |
| ASL | American Sign Language |
| ISL | Indian Sign Language |
| SLR | Sign Language Recognition |
| CSLR | Continuous Sign Language Recognition |
| LSTM | Long Short-Term Memory |
| VoIP | Voice over Internet Protocol |
| KNN | K-Nearest Neighbour |
| ANN | Artificial Neural Network |
| SVM | Support Vector Machine |
| HMM | Hidden Markov Model |
| CNN | Convolutional Neural Network |
| RNN | Recurrent Neural Network |
| LMC | Leap Motion Controller |
| CRF | Conditional Random Fields |
| 3D-ResNet | 3D residual network |
| CTC | Connectionist Temporal Classification |

**DECLARATIONS**

**Authors' contributions:** Conceptualization: Sharvani Srivastava, Sudhakar Sing, Shiv Prakash; Methodology: Sharvani Srivastava, Sudhakar Sing, Shiv Prakash; Formal analysis and investigation: Sharvani Srivastava, Sudhakar Singh, Pooja, Shiv Prakash; Writing - original draft preparation: Sharvani Srivastava, Sudhakar Singh; Writing - review and editing: Sudhakar Singh, Pooja, Shiv Prakash; Resources: Sharvani Srivastava, Sudhakar Singh, Pooja, Shiv Prakash; Supervision: Sudhakar Singh, Pooja, Shiv Prakash.

**Conflicts of interest:** The authors declare that they have no conflict of interest in this paper.

**Funding:** No funding was received for conducting this study.

**Availability of data and material:** Not applicable.

**Code availability:** Code available on request from the authors.